\PassOptionsToPackage{table}{xcolor}
\documentclass[10pt,twocolumn,letterpaper]{article}

\usepackage[pagenumbers]{cvpr} 

\definecolor{cvprblue}{rgb}{0.21,0.49,0.74}
\usepackage[pagebackref,breaklinks,colorlinks,citecolor=cvprblue]{hyperref}
\usepackage{algorithm}
\usepackage{algorithmic}
\usepackage{graphicx}
\usepackage{amsfonts}
\usepackage{multirow}
\usepackage{amssymb}
\usepackage{newtxmath}
\usepackage{url}
\usepackage{bm}
\usepackage{marvosym}
\usepackage{enumitem}
\usepackage{booktabs}       
\usepackage{nicefrac}       
\usepackage{microtype}      
\usepackage{multicol}
\usepackage{amsmath}
\usepackage{mathrsfs}
\usepackage{algorithm}
\usepackage{algorithmic}
\usepackage{ulem}
\usepackage{amssymb}
\usepackage{threeparttable}
\usepackage[accsupp]{axessibility}

\def\paperID{4490} 
\def\confName{CVPR}
\def\confYear{2025}

\title{Efficient Transfer Learning for Video-language Foundation Models }

\author{Haoxing Chen$^{1}$, Zizheng Huang$^{1,2,3}$, Yan Hong$^{1}$, Yanshuo Wang$^{1,4}$, Zhongcai Lyu$^{1}$, \\Zhuoer Xu$^{1}$, Jun Lan$^{1}$, Zhangxuan Gu$^{1}$\thanks{Corresponding author. Code will be avaliable at  \url{https://github.com/chenhaoxing/ETL4Video}.}\\
$^{1}$Tiansuan Lab, Ant Group,
$^{2}$Nanjing Univeristy\\
$^{3}$Shanghai Innovation Institute,
$^{4}$Australian National University\\
{\tt\small hx.chen@hotmail.com}
}

\begin{document}
\maketitle

\begin{abstract}
Pre-trained vision-language models provide a robust foundation for efficient transfer learning across various downstream tasks. In the field of video action recognition, mainstream approaches often introduce additional modules to capture temporal information. 
Although the additional modules increase the capacity of model, enabling it to better capture video-specific inductive biases, existing methods typically introduce a substantial number of new parameters and are prone to catastrophic forgetting of previously acquired generalizable knowledge.
In this paper, we propose a parameter-efficient Multi-modal Spatio-Temporal Adapter (MSTA) to enhance the alignment between textual and visual representations, achieving a balance between generalizable knowledge and task-specific adaptation.
Furthermore, to mitigate over-fitting and enhance generalizability, we introduce a spatio-temporal description-guided consistency constraint. 
This constraint involves providing template inputs (e.g., "a video of \{\textbf{cls}\}") to the trainable language branch and LLM-generated spatio-temporal descriptions to the pre-trained language branch, enforcing output consistency between the branches. This approach reduces overfitting to downstream tasks and enhances the distinguishability of the trainable branch within the spatio-temporal semantic space.
We evaluate the effectiveness of our approach across four tasks: zero-shot transfer, few-shot learning, base-to-novel generalization, and fully-supervised learning. Compared to many state-of-the-art methods, our MSTA achieves outstanding performance across all evaluations, while using only 2-7\% of the trainable parameters in the original model. 
\end{abstract}

\section{Introduction}

\begin{figure}[t]
\centering
\includegraphics[width=0.99\linewidth]{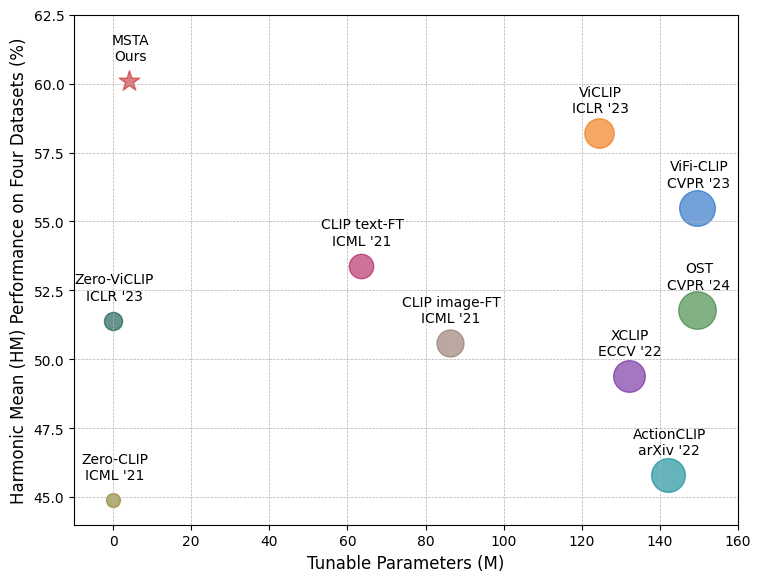}
\caption{We compared the number of trainable parameters of our method with other state-of-the-art methods, as well as the average Harmonic Mean Performance on the base-to-novel generalization task across four datasets. It can be observed that our method uses fewer parameters while achieving state-of-the-art results.}
	\label{tunable_params}
\end{figure}

Multi-modal foundation models (e.g., CLIP~\cite{CLIP}, ViCLIP~\cite{InternVid}, BLIP2~\cite{BLIP2}), trained on large-scale paired multi-modal datasets, have demonstrated exceptional generalization capabilities. The semantic visual concepts enabled by extensive multi-modal alignment can be effectively transferred to various downstream tasks, such as few-shot image/video classification~\cite{MaPLe,CPR,Tip-Adapter,ViFiCLIP}, open-vocabulary detection~\cite{PromptDet,CORA}, and segmentation~\cite{CLIPSeg}. In this paper, we focus on transferring the pre-trained multi-modal foundation models model (i.e., CLIP~\cite{CLIP} and ViCLIP~\cite{InternVid}) to video action recognition tasks, highlighting their significant potential to advance research in this domain.

The key to the adaptation process of foundational multi-modal models lies in injecting domain-specific expertise. In video action recognition, this need is particularly evident due to the dynamic and temporal nature of video data, requiring models to effectively capture context and temporal relationships. Consequently, existing methods typically enhance pre-trained CLIP models by integrating carefully designed prompts~\cite{Vita-CLIP}, adapters~\cite{MoTE,ST-Adapter,AIM}, or temporal modules~\cite{XCLIP} to better acquire video-specific knowledge. For instance, XCLIP~\cite{XCLIP} employs cross-frame attention and multi-frame integration modules to model temporal information. ActionCLIP~\cite{ActionCLIP} uses pre-training to generate robust representations, applying prompt engineering to align the action recognition task closely with the pre-trained model, followed by fine-tuning to achieve superior performance. Despite the increased model capacity resulting from numerous learnable parameters, such approaches can lead to catastrophic forgetting, thereby diminishing the generalizability of the original models.

Alternatively, Wang \textit{et al.}~\cite{InternVid} proposed ViCLIP to enhance CLIP performance specifically for video understanding. ViCLIP is trained on extensive video-text pairs and incorporates spatiotemporal attention within the visual encoder, retaining other CLIP components unchanged. However, current research has not yet explored effective transfer learning methods specifically designed for ViCLIP, and existing CLIP-based methods cannot be directly applied.

Thus, the critical question arises: \textit{Is there an efficient fine-tuning method for pre-trained models that preserves generalizability?} One possible approach is directly applying efficient transfer methods used for image/text foundation models, such as LoRA~\cite{LoRA} and AdaptFormer~\cite{AdaptFormer}, to video tasks. However, these methods primarily rely on uni-modal information, making them unsuitable for multi-modal models. Another simple strategy involves independently applying adapters to each modality, but this overlooks the interrelation between text and video representations. Consequently, directly applying identical adapters may not effectively capture task-specific nuances that vary across vision and language modalities. Furthermore, existing approaches do not fully address the distinct characteristics inherent in text and video representations. Effective transfer learning typically involves balancing discrimination and generalization—fine-tuning task-specific discriminative features while preserving features broadly applicable across different tasks.

To this end, we propose a novel Mutli-modal Spatio-Temporal Adapter (MSTA) architecture for multi-modal foundation models such that the text and video representations can be better aligned. As shown in Fig.~\ref{tunable_params}, our method achieves the best generalization performance with a minimal number of parameters. Specifically, our MSTA includes independent projection layers for the text and video branches to learn task-specific knowledge for each modality. To achieve effective alignment between modalities, we introduce a shared unified feature projection layer, which is jointly utilized by both modalities. During the fine-tuning phase, this shared feature space can receive gradients from both modalities~\cite{MMA}, thereby optimizing the alignment between them. For the video branch, we further design two simple up-projection layers to enhance the adaptation capabilities for temporal and spatial features. To further mitigate over-fitting and enhance the generalizability, we introduce a spatio-temporal description-guided consistency constraint. This method successfully transfers knowledge from a frozen encoder to a learnable encoder through knowledge distillation, allowing the model to maintain the generalization strength of the pre-trained base model when handling new tasks in few-shot scenarios~\cite{CoPrompt}. Specifically, we impose a consistency constraint on the text branch between the trainable model with MSTA and the pre-trained model. We leverage a pre-trained large language model (e.g., DeepSeek~\cite{DeepSeek-V2}) to generate more detailed and descriptive sentences of temporal and spatial features, and impose a consistency constraint between the representations of the learnable text encoder and the pre-trained text encoder based on these sentences. To validate the effectiveness of our model, we perform extensive experiments across six benchmark datasets: Kinetics-400~\cite{K400}, Kinetics-600~\cite{K600}, UCF-101~\cite{UCF101}, HMDB-51~\cite{HMDB}, SomethingSomething V2~\cite{ssvth}, and ActivityNet~\cite{ActivityNet}. The results show that our approach attains state-of-the-art performance in open-vocabulary tasks, such as zero-shot and few-shot learning, and consistently enhances performance when integrated with existing pipelines in fully supervised settings. 

The key contributions of this work are as follows:
\begin{itemize}
    \item We propose a novel Multi-modal Spatio-Temporal Adapter (MSTA) that contains separate projection layers to improve feature representations for video and text encoders independently. Additionally, we implement a shared projection layer to improve the alignment between video and language representations.
    \item We propose a spatio-temporal description-guided consistency constraint for large multi-modal foundation models, enabling them to learn new tasks from a small number of samples while maintaining their generalization capability.
    \item Extensive experiments demonstrate that MSTA achieves an optimal balance between new and old knowledge while training only a small number of parameters. Comprehensive ablation studies showcase the scalability and effectiveness of our proposed method.
\end{itemize}

\section{Related Works}
\textbf{Multi-modal Foundation Models.}
The latest advances in multi-modal foundation models have significantly impacted the field of computer vision, especially in tasks that combine language and vision. Representative models include, but are not limited to, CLIP~\cite{CLIP}, BLIP~\cite{BLIP2}, Florence~\cite{Florence}, Kosmos~\cite{Kosmos2}, InternVideo~\cite{InternVideo}, and ViCLIP~\cite{InternVid}. These models leverage self-supervised paradigms extracted from large-scale multi-modal web data for training. For example, CLIP~\cite{CLIP} is trained using contrastive loss~\cite{contras_loss} on approximately 400 million image-text pairs, while ViCLIP~\cite{InternVid} is trained on about 10 million video-text pairs. By collecting more multi-modal data, these models have demonstrated promising performance across various downstream applications. Despite their ability to learn generalized representations, effectively adapting these pre-trained models to specific downstream tasks remains a major challenge, particularly in few-shot settings. To address this, many studies have proposed various methods tailored to different tasks, such as few-shot action recognition~\cite{ActionCLIP,ViFiCLIP}, video question answering~\cite{ZFL_VQA}, and segmentation~\cite{DZSL}. In contrast, this work proposes a novel multi-modal spatio-temporal adapter to effectively adapt multi-modal foundation models for generalization tasks.

\noindent
\textbf{Efficient Transfer Learning.} Traditional approaches~\cite{bert,ViT} migrate models to downstream tasks by fine-tuning all parameters of the pre-trained network. However, as model sizes increase, this traditional paradigm faces a significant computational burden, and fine-tuning a large number of parameters often leads to severe overfitting, especially in low-sample scenarios. Recently, numerous methods~\cite{MaPLe,CPR,Tip-Adapter,ViFiCLIP} have been proposed to explore the adaptation of pre-trained vision-language models~\cite{CLIP,InternVid} to downstream tasks. In this paper, we focus on transferring pre-trained models to video understanding tasks. ViFi-CLIP~\cite{ViFiCLIP} demonstrates that direct fine-tuning exhibits good generalization capabilities across various settings. Open-VCLIP~\cite{Open-VCLIP} constructs an open-vocabulary video model by interpolating the model weights and its optimization trajectory. Vita-CLIP~\cite{Vita-CLIP} uses multi-level prompts to extract discriminative information. X-CLIP~\cite{XCLIP} introduces cross-frame attention and multi-frame integration modules for temporal modeling. OST~\cite{OST} optimizes text knowledge by leveraging large language models to generate spatio-temporal descriptors, and proposes an optimal descriptor solver to enhance generalization. MoTE~\cite{MoTE} inserts a mixture-of-temporal-experts into the visual encoder to capture diverse views of data bias. However, all of these methods require learning a large number of parameters and suffer from catastrophic forgetting of the original generalizable knowledge. Moreover, these methods are specifically designed for CLIP~\cite{CLIP}, making them unsuitable for the latest base model, i.e., ViCLIP~\cite{InternVid}. ViCLIP is a pre-trained model based on CLIP, trained on 10 million video-text pairs. Our work addresses these challenges by designing an efficient tranfer leraning approach tailored for ViCLIP, demonstrating superior results across various settings.

\section{Methods}
\subsection{Preliminaries}
We adopt ViCLIP~\cite{InternVid} as the pre-trained video-language foundation model in our method. ViCLIP is an improved version of CLIP that replaces the native attention in the visual encoder with spatiotemporal attention, while keeping other design components unchanged. It uses pre-trained CLIP weights for initialization. The model is trained on the InternVid-10M dataset~\cite{InternVid}, with the optimization objective being the contrastive loss between input video and text embeddings. ViCLIP is composed of two branches: a text branch with encoder $E_t$ and a vision branch with encoder $E_v$. These two branches enable it to understand and bridge the semantic gap between textual descriptions and visual content. In particular, a video $V \in \mathbb{R}^{T\times H\times W\times3}$ will be fed into the encoder $E_v$ to obtain the video feature $x$ as follows:
\begin{align}
    x_0 &= \text{PatchEmbed}(V), \label{visualembed}\\
    [c_i,x_i] &= \mathcal{E}^i_v([c_{i-1}, x_{i-1}]),\,i=1,2,...,L,\label{each_visual_layer}\\
    x &= \text{PatchProj}(c_L).
\end{align}
Here, $\text{PatchEmbed}$ first divides the input video $V$ into fixed-size patches, projecting them into feature embeddings. A learnable class token $c_0$ is then concatenated with these embeddings, forming $[c_0, x_0]$, which is subsequently passed through $L$ transformer blocks, represented as $\{\mathcal{E}^i_v\}_{i=1}^{L}$. Finally, the class token $ c_L$ from the last transformer block $\mathcal{E}^L_v$ is projected into the video feature $ x $ via a projection layer $\text{PatchProj}$, positioning the feature in the shared vision-language space. In a similar manner, a text description $T$ is processed through the text encoder $E_t$ to obtain the text feature $w$ as follows:
\begin{align}
w_0 &= \text{TextEmbed}(T),\\
w_i &= \mathcal{E}^i_t(w_{i-1}),\, i=1,2,...,L,\label{each_text_layer}\\
w &= \text{TextProj}(w_L).
\end{align}
As shown, this process involves three steps: first, $\text{TextEmbed}$ is used to tokenize and project the input text description into the initial word embedding $w_0$. Next, a series of transformer blocks $ \{\mathcal{E}_t^i\}_{i=1}^{L}$ progressively abstracts the features, producing a refined embedding $ w_L $ at the final layer. Finally, $\text{TextProj}$ projects the output $ w_L $ from the last transformer block $ \mathcal{E}_t^L$ into the common vision-language space. Given these features, we compute the cosine similarity scores $sim(x, w) $ between the video and text descriptions across different domains or tasks to perform task-specific predictions.

\begin{figure}[t]
	\centering
	\includegraphics[width=0.99\linewidth]{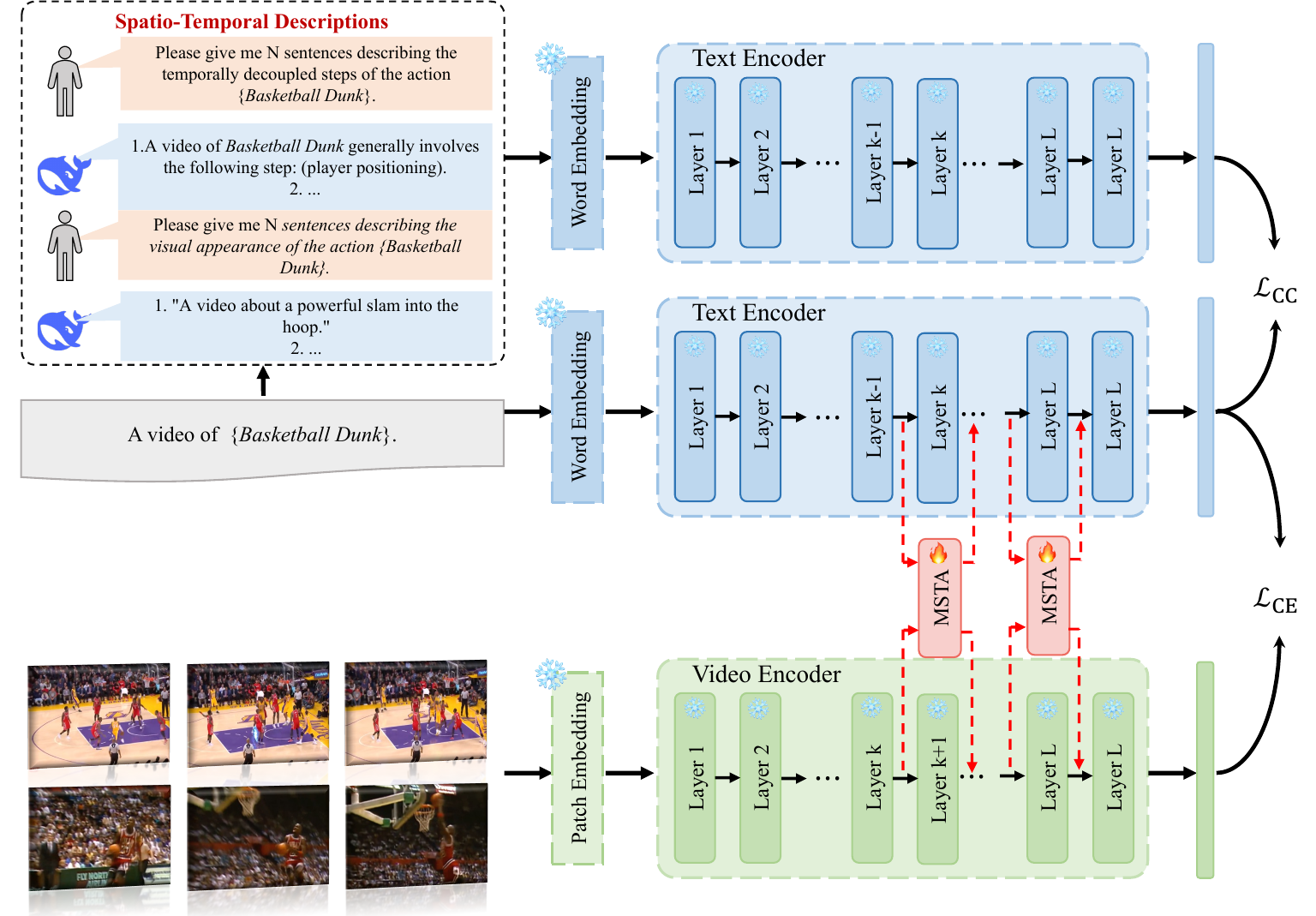}
	\caption{The proposed Multi-modal Spatio-Temporal Adapter (MSTA) is designed for transformer-based multi-modal foundation models. It optimizes only the additional adapters, keeping the pretrained CLIP model frozen. To balance discrimination and generalization, we selectively fine-tune a few higher layers ($\geq k$). MSTA employs shared weights for video and text representations, enabling the model to capture shared cues from both modalities.}
	\label{MAIN_ARC}
\end{figure}

\subsection{Multi-modal Spatio-Temporal Adapter}
Our work mainly focuses on generalization tasks (i.e., few-shot generalization~\cite{Tip-Adapter,MaPLe} and zero-shot generalization~\cite{ActionCLIP}), where the pre-trained multi-modal foundation models are first fine-tuned on a limited set of training samples, and then applied to recognize unseen instances. For such problems, a good instance representation should not only be discriminative but also exhibit strong generalization capabilities across different types of datasets. Typically, adding additional parameters helps better fit the training data distribution, thereby improving performance on seen tasks. However, models optimized for specific target distributions are often highly sensitive to external distribution shifts, which may lead to poor generalization performance in downstream tasks involving unknown video categories. While, in theory, expanding the training dataset to cover more unseen categories could mitigate this issue, the high computational and data acquisition costs associated with video data make this approach impractical. Therefore, it is crucial to develop a foundational model architecture with enhanced generalization capability. 

Recently, parameter efficient fine tuning~\cite{LoRA,AdaptFormer} (PEFT) has proven to be an effective method for transfer learning in large models. Two representative approaches are LoRA~\cite{LoRA} and AdaptFormer~\cite{AdaptFormer}. However, these methods are designed for uni-modal models and cannot be directly applied to multi-modal models. While it is possible to apply adapters separately to the two modalities, the lack of interaction between visual and textual information limits their performance. Therefore, we propose a new adapter-based efficient transfer learning framework as described below.

\textbf{Holistic design.}  As illustrated in Fig.~\ref{MAIN_ARC}, unlike most existing methods that introduce adapters or tokens across the entire network or into lower layers, our adapter $\mathcal{A}$ is selectively incorporated into only a few higher layers of both the video and text encoders. Specifically, for the video encoder $E_v$, we incorporate adapters $\{\mathcal{A}_v^j\}_{j=k}^{L}$ starting from the $k$-th transformer block, modifying Eq.~(\ref{each_visual_layer}) as follows:
\begin{align}
    [c_i,x_i] &= \mathcal{E}^i_v([c_{i-1}, x_{i-1}]),\, i=1,2,\dots,k-1, \\
    [c_j,x_j] &= \mathcal{E}^j_v([c_{j-1}, x_{j-1}]) + \lambda \boxed{\mathcal{A}^j_v([c_{j-1}, x_{j-1}])},\notag\\ j&=k,k+1,\dots,L. 
\end{align}
Here, the portion inside the box represents the trainable blocks. The coefficient $\lambda$ is used to balance task-specific knowledge with general pre-trained knowledge. Notably, setting $\lambda=0$ reduces the model to the original transformer block, without incorporating any additional knowledge. Similarly, we add adapters $\{\mathcal{A}_t^j\}_{j=k}^{L}$ to the text encoder $E_t$ and modified Eq.~(\ref{each_text_layer}) as follows:
\begin{align}
    w_i &= \mathcal{E}^i_t(w_{i-1}),\quad i=1,2,\dots,k-1,\\
    w_j &= \mathcal{E}^j_t(w_{j-1}) + \lambda \boxed{\mathcal{A}^j_t(w_{j-1})},\, j=k,k+1,\dots,L. 
\end{align}

\begin{figure}[t]
	\centering
	\includegraphics[width=0.9\linewidth]{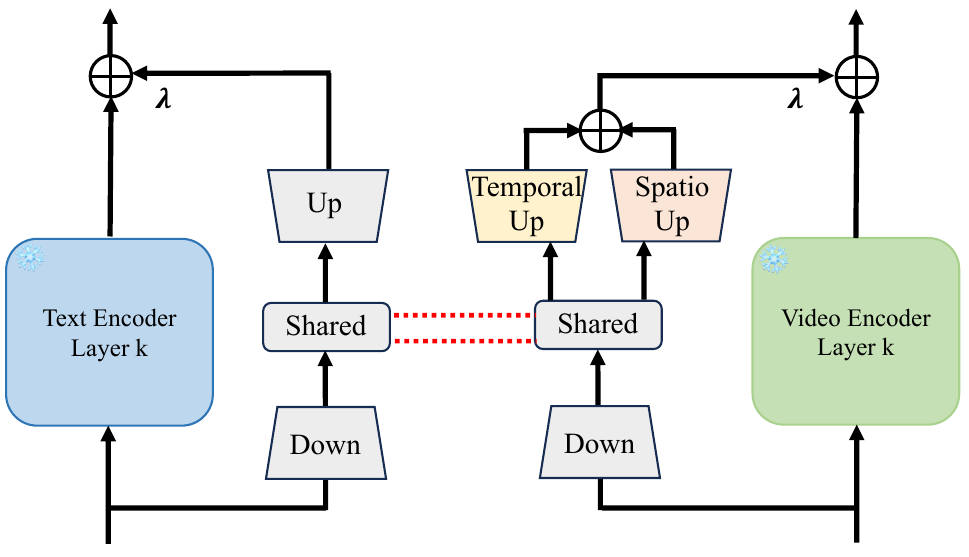}
	\caption{Overview of the proposed Multi-modal Spatio-Temporal Adapter (MSTA).}
	\label{MSTA}
\end{figure}

\textbf{Specific design.} 
According to ~\cite{Adapter}, the adapters in both the text branch and the video branch consist of down-projection layers (encoders), adapter layers, and up-projection layers (decoders). As shown in Fig.~\ref{MSTA}, to bridge the representations in both branches and reduce the semantic gap, we do not independently adapt the adapters in the video and text branches. Instead, we aggregate these bimodal signals through a shared projection layer. During fine-tuning, this shared feature space receives gradients from both modalities, thereby optimizing their alignment. Additionally, for the video branch, we designed two types of up-projection layers: one spatial up-projection layer and one temporal up-projection layer, to further enhance the adaptability to spatiotemporal features. Formally,
this process can be summarized as follows:
\begin{align}
    \mathcal{A}^k_v(h_k) &= \mathbf{W}_v^{ku-s} \cdot h_k+ \mathbf{W}_v^{ku-t} \cdot h_k, \\
    h_k &= \sigma(\mathbf{W}^{ks}\cdot \sigma(\mathbf{W}_v^{kd}\cdot [c_k,x_k])).
\end{align}
A similar process is added to text encoder as follows:
\begin{equation}
    \mathcal{A}^k_t(w_k) = \mathbf{W}_t^{ku}\cdot \sigma(\mathbf{W}^{ks}\cdot \sigma(\mathbf{W}_t^{kd}\cdot w_k)).
\end{equation}
Here, $\mathbf{W}^{ku}$ and $\mathbf{W}^{kd}$ represent the "Up" and "Down" projection layers of the $k$-th layer, as illustrated in Fig.~\ref{MSTA}, with the modality branch indicated by the subscripts. $\mathbf{W}_v^{ku-s}$ is the spatial up-projection layer implemented by a linear layer, while $\mathbf{W}_v^{ku-t}$ is the temporal up-projection layer implemented by a 3D convolution layer. $\mathbf{W}^{ks}$ denotes the $k$-th projection layer, which is shared across different branches. 

\subsection{Spatio-Temporal Description-guided Consistency Constraint}
To address the issue of decreased generalization ability due to over-fitting on downstream tasks, we propose the Spatio-Temporal Description-guided Consistency constraint. This constraint ensures that the embeddings generated by the trainable model (adjustable adapter parameters in the image and text branches) do not significantly deviate from those generated by the pre-trained multi-modal foundation model, while enhancing the understanding of temporal and spatial features. In the language branch, we leverage a pre-trained large language model (LLM)~\cite{DeepSeek-V2} to generate more descriptive sentences that better capture the temporal and spatio aspects of actions. The spatio description aims to capture static visual elements that can be discerned from a single image, such as scenes and common objects. The temporal description, on the other hand, is designed to decompose action categories in a step-by-step manner, describing the temporal evolution of actions. To generate spatio descriptions,  we use the following prompt with category name $\{\textbf{cls}\}$ to query LLM: \textcolor{gray}{``\textit{Please give me N sentences describing the visual appearance of the action $\{\textbf{cls}\}$."}}. For temporal descriptions, we utilize the the following temporal prompt: \textcolor{gray}{``\textit{Please give me N sentences describing the temporally decoupled steps of the action $\{\textbf{cls}\}$."}}. Through the above operations, we obtain the spatio descriptions $\text{DES}_s$ and temporal $\text{DES}_t$, each containing $N$ descriptions.

Subsequently, we feed sentences based on standard template (i.e., a video of $\{\textbf{cls}\}$.) into the learnable branch, while spatio/temporal descriptions are provided to the pre-trained branch. We employ cosine distance as a consistency constraint between the embeddings from the pre-trained branch and the learnable branch. The pre-trained encoder extracts features from the spatio/temporal descriptions, allowing the consistency constraint to be defined as:
\begin{equation}
    \mathcal{L}_{\rm CC} = 2 - \frac{w^c\cdot D^c_s}{||w^c\cdot D^c_t||||D^c_s||} - \frac{w^c}{||w^c||||D^c_t||},
\end{equation}
where $w^c$ indicates the text embedding for class $c$ of the pre-trained branch, and $D^c_s$ represents the mean embedding of the spatio descriptions for class $c$, obtained by averaging the features of the input descriptions processed through the pre-trained branch. Similarly, $D^c_t$ is computed for the temporal descriptions. This consistency constraint loss is combined with a supervised loss to create the final loss function. The supervised loss is given by:
\begin{equation}
    \mathcal{L}_{\rm CE} = -log \frac{{\rm exp}(sim(x, w^c)/\tau}{\sum_{k=1}^C{\rm exp}(sim(x, w^k)/\tau},
\end{equation}
where $\tau$ is a temperature parameter controlling the sharpness of similarity scores. By combining these losses with a weighting factor $\alpha$, the final loss function for MSTA becomes:
\begin{equation}
    \mathcal{L} = \mathcal{L}_{\rm CE} + \alpha\mathcal{L}_{\rm CC}.
\end{equation}

\begin{table}[!t]
\centering
\caption{Implementation details of MSTA}
\resizebox{\linewidth}{!}{
\begin{tabular}{c|ccccccccc}
\toprule
Dataset & Batch size & Learning rate & Training epochs & $\alpha$ & $\lambda$ & Dims & $N$ & Dropout & Layers \\
\midrule
\rowcolor{blue!5}\multicolumn{10}{c}{Base-to-novel generalization} \\
K-400 & 32 & 1.e-03 & 11 & 1.0 & 0.005 & 256 & 2 & 0.1 & 1-12 \\
SSv2 & 32 & 1.e-03 & 11 & 1.0 & 0.005 & 256 & 2 & 0.1 & 1-12 \\
HMDB-51 & 32 & 1.e-03 & 11 & 1.0 & 0.005 & 256 & 2 & 0.1 & 1-12 \\
UCF-101 & 32 & 1.e-03 & 11 & 1.0 & 0.005 & 256 & 2 & 0.1 & 1-12 \\
\midrule
\rowcolor{orange!5}\multicolumn{10}{c}{Few-shot learning} \\
SSv2 & 32 & 1.e-03 & 50 & 1.0 & 0.005 & 128 & 2 & 0.1 & 8-12 \\
HMDB-51 & 32 & 1.e-03 & 50 & 1.0 & 0.005 & 128 & 2 & 0.1 & 8-12 \\
UCF-101 & 32 & 1.e-03 & 50 & 1.0 & 0.005 & 128 & 2 & 0.1 & 8-12 \\
\midrule
\rowcolor{yellow!5}\multicolumn{10}{c}{Zero-shot Transfer} \\
K-400 & 256 & 8.e-03 & 100 & 1.0 & 0.001 & 128 & 2 & 0.1 & 1-12 \\
\midrule
\rowcolor{green!5}\multicolumn{10}{c}{Few-shot learning} \\
K-400 & 256 & 8.e-03 & 100 & 1.0 & 0.001 & 128 & 2 & 0.1 & 1-12 \\
\bottomrule
\end{tabular}}
\label{Implementation}
\end{table}

\begin{table*}[t]
    \centering
    \caption{Base-to-novel generalization: We compare the generalization ability of MSTA with models that adapt ViCLIP for video tasks. “HM” is the harmonic mean of base and novel accuracy, providing the trade-off between adaption and generalization. "Zero-ViCLIP" refers to zero-shot ViCLIP, which directly uses the pre-trained ViCLIP for inference.}
    \resizebox{\textwidth}{!}{
    \begin{tabular}{cccccccccccccc}
        \toprule
        \multirow{2}{*}{Method}&\multirow{2}{*}{GFLOPs}& \multicolumn{3}{c}{K-400} & \multicolumn{3}{c}{HMDB-51} & \multicolumn{3}{c}{UCF-101} & \multicolumn{3}{c}{SSv2} \\
        \cmidrule(lr){3-5} \cmidrule(lr){6-8} \cmidrule(lr){9-11} \cmidrule(lr){12-14}
        & & Base & Novel & HM & Base & Novel & HM & Base & Novel & HM & Base & Novel & HM \\
        \midrule
        \rowcolor{gray!5}\multicolumn{13}{c}{Zero-shot pre-trained models}\\
        Vanilla CLIP$_{\textcolor[rgb]{0.5, 0.5, 0.5}{\rm (ICML'21)}}$&281 &55.1 & 55.2&55.1 &50.6&48.1 &49.3 &78.0&63.8 &70.2 &4.8&5.4&5.1\\
        \midrule
        \rowcolor{yellow!5}\multicolumn{13}{c}{Adapting pre-trained models}\\
        ActionCLIP$_{\textcolor[rgb]{0.5, 0.5, 0.5}{\rm (arXiv'22)}}$&282 &  61.5& 47.2&53.4 & 69.3& 38.3& 49.3 & 90.5& 58.7& 71.2 &13.5& 10.4& 11.7 \\
        XCLIP$_{\textcolor[rgb]{0.5, 0.5, 0.5}{\rm (ECCV'22)}}$ & 145&74.7& 56.7& 64.5& 69.9& 45.3& 55.0& 90.7& 59.5& 71.9 & 8.9& 6.8&7.7 \\
        A5$_{\textcolor[rgb]{0.5, 0.5, 0.5}{\rm (ECCV'22)}}$&284 & 69.9& 38.4& 49.6 & 46.9& 16.5&24.4 & 90.9& 40.8& 56.3 &8.6& 6.4 &7.3 \\

        \midrule
        \rowcolor{green!5}\multicolumn{13}{c}{Tuning pre-trained models}\\
        Vanilla CLIP$_{\textcolor[rgb]{0.5, 0.5, 0.5}{\rm (ICML'21)}}$&281 &76.4 &63.0 &69.1 &71.5 &55.3&62.4 &92.3 &67.9 &78.2 &13.2&10.8&11.9\\ 
        ViFi-CLIP$_{\textcolor[rgb]{0.5, 0.5, 0.5}{\rm (CVPR'23)}}$ &281 &76.8& 61.3&68.2 & 74.5& 53.9&62.5 & 92.5& 68.5& 78.7& 16.5& 12.4 &14.2\\
        OST$_{\textcolor[rgb]{0.5, 0.5, 0.5}{\rm (CVPR'24)}}$&- &75.4 &59.6 &66.6 &75.1 &34.4 &47.2 & \textbf{96.1}&68.9&80.3&15.3&11.5&13.1\\
        ViCLIP$_{\textcolor[rgb]{0.5, 0.5, 0.5}{\rm (ICLR'23)}}$ & 161&\textbf{78.5} & 65.3 &71.3 &77.9 &52.4& 62.7&95.1& 71.5&81.6&19.4& 15.2&17.0\\
        \midrule
        \rowcolor{blue!5}\multicolumn{13}{c}{Efficient transfer pre-trained models}\\
         Zero-ViCLIP$_{\textcolor[rgb]{0.5, 0.5, 0.5}{\rm (ICLR'23)}}$&161.74 & 68.7&63.4 &65.9 &62.5 &48.9 &54.9 &81.7 &69.8  &75.3 &9.3 &8.9&9.1 \\
         + AdaptFormer$_{\textcolor[rgb]{0.5, 0.5, 0.5}{\rm (NeurIPS'22)}}$&179.80 &77.7 & 65.6&71.1 &\textbf{77.5}  & 54.9 &64.3 & 94.9 & 72.7 &82.3 &20.4& 14.6   &17.0 \\
         + LoRA$_{\textcolor[rgb]{0.5, 0.5, 0.5}{\rm (NeurIPS'22)}}$&179.80 &77.6&65.3&70.9&\uline{77.2} &54.6 &64.0& 95.1&72.3 &82.1&19.5 &13.6 & 16.0\\
         + MSTA& 179.84&\uline{78.2}&\uline{66.2}&\uline{71.7}&\uline{77.2}&\uline{57.5}&\uline{65.9}&\uline{95.6} &\uline{72.8} & \uline{82.7}&\uline{21.9}& \uline{16.3}&\uline{18.7} \\
         + MSTA + $\mathcal{L}_{\rm CC}$&179.84 &\textbf{78.5} 
&\textbf{66.5} & \textbf{72.0} &\textbf{77.5} & \textbf{57.9} & \textbf{66.3} & \uline{96.0}&\textbf{72.9}  & \textbf{82.9}&\textbf{22.2}&\textbf{16.5} &\textbf{18.9} \\
        \bottomrule
    \end{tabular}}
    \label{base2new}
\end{table*}

\begin{table*}[t]
\centering
\caption{Comparisons with state-of-the-art methods for few-shot video recognition on HMDB51, UCF101 and Something-Something V2. We scaled up the task to categorize all categories in the dataset with only a few samples per category for training. Here $K$ denotes training samples for each class. We report Top-1 accuracy using multi-view inference.}
\resizebox{\textwidth}{!}{
\begin{tabular}{cccccccccccccc}
\toprule
\multirow{2}{*}{Method}&\multirow{2}{*}{GFLOPs} & \multicolumn{4}{c}{HMDB-51}& \multicolumn{4}{c}{UCF-101}& \multicolumn{4}{c}{SSv2} \\ \cmidrule(lr){3-6} \cmidrule(lr){7-10} \cmidrule(lr){11-14}
&&$K=2$&$K=4$&$K=8$&$K=16$&$K=2$&$K=4$&$K=8$&$K=16$&$K=2$&$K=4$&$K=8$&$K=16$\\ \midrule
\rowcolor{gray!10}\multicolumn{13}{c}{Zero-shot pre-trained models}\\
Vanilla CLIP$_{\textcolor[rgb]{0.5, 0.5, 0.5}{\rm (ICML'21)}}$&281 & 37.2 &37.2 &37.2 &37.2 &62.8 &62.8 &62.8 &62.8 &2.8 &2.8 &2.8 &2.8\\ 
\midrule
\rowcolor{yellow!5}\multicolumn{13}{c}{Adapting pre-trained models}\\
XCLIP$_{\textcolor[rgb]{0.5, 0.5, 0.5}{\rm (ECCV'22)}}$&145  &53.5& 57.8& 62.9& 64.5& 49.3& 76.2& 84.1& 91.6& 4.4& 5.2& 6.6& 11.1 \\ 
A5$_{\textcolor[rgb]{0.5, 0.5, 0.5}{\rm (ECCV'22)}}$&284  &41.2& 51.3& 56.5& 62.7& 71.7& 80.3& 85.9& 90.3& 4.5& 5.3& 6.3& 9.9 \\ 
\midrule
\rowcolor{green!5}\multicolumn{13}{c}{Tuning pre-trained models}\\
Vanilla CLIP$_{\textcolor[rgb]{0.5, 0.5, 0.5}{\rm (ICML'21)}}$& 282&57.8 &61.2 &65.6 &66.7 &81.8 &86.3 &90.1 &92.5 &6.7 & 7.3&9.0 &12.2\\ 
ViFi-CLIP$_{\textcolor[rgb]{0.5, 0.5, 0.5}{\rm (CVPR'23)}}$&281 & 57.5& \uline{62.9}& 64.7 &66.6& 80.5& 85.3& 90.2& 93.2 &6.5 &7.7&8.3 &12.9\\ 
OST$_{\textcolor[rgb]{0.5, 0.5, 0.5}{\rm (CVPR'24)}}$&- &60.0&{62.5} &65.6 &67.3 & 83.0&88.4 &91.3 &\uline{93.9} &7.3&8.4 &8.5 &11.5\\
MoTE$_{\textcolor[rgb]{0.5, 0.5, 0.5}{\rm (NeurIPS'24)}}$& 141&\uline{61.0} &\textbf{63.7} & \uline{66.9} & 68.5 &\textbf{88.1} & \textbf{90.7} &{92.2} &{93.7} &7.4 &8.7 &9.8 &12.5\\
ViCLIP$_{\textcolor[rgb]{0.5, 0.5, 0.5}{\rm (ICLR'23)}}$&161 &53.7& 60.4 &64.5& \uline{70.3} &83.0 &88.0 &92.1 &93.2 &{8.7} &{9.7} &{11.6} &{15.4} \\
\midrule
\rowcolor{blue!5}\multicolumn{13}{c}{Efficient transfer pre-trained models}\\
Zero-ViCLIP$_{\textcolor[rgb]{0.5, 0.5, 0.5}{\rm (ICLR'23)}}$&161 &47.8 &47.8 &47.8 &47.8 &71.0 &71.0 &71.0 &71.0  & 5.1& 5.1& 5.1&5.1 \\
+ AdaptFormer$_{\textcolor[rgb]{0.5, 0.5, 0.5}{\rm (NeurIPS'22)}}$ &179.80 &60.0 & 61.8&66.6 &70.1 &84.8&88.7 &91.7 &93.3 &8.5 &10.1 &11.5 &15.2 \\
+ LoRA$_{\textcolor[rgb]{0.5, 0.5, 0.5}{\rm (NeurIPS'22)}}$&179.80 &59.4 & 61.4&66.5 &69.4 &83.5&88.3 &90.9&92.7 &8.4 &9.5 &11.1 &14.9 \\
+ MSTA&179.84 &60.1&62.2&66.8 &69.9&85.1&89.0&\uline{92.3}&93.5&\uline{9.0}&\uline{10.1}&\uline{12.7}&\uline{16.6} \\
+ MSTA + $\mathcal{L}_{\rm CC}$&179.84  &\textbf{61.2} &62.3 &\textbf{67.1} &\textbf{70.4} &\uline{86.1}  &\uline{90.2} &\textbf{92.7} &\textbf{94.8} &\textbf{9.1}&\textbf{10.4}&\textbf{13.4} &\textbf{17.5}  \\
\bottomrule
\end{tabular}}
\label{fewshot}
\end{table*}

\section{Experiments}






\subsection{Experimental Setup}
\textbf{Dataset.} We conducted experiments on six video benchmarks: Kinetics-400~\cite{K400}, Kinetics-600~\cite{K600}, UCF-101~\cite{UCF101},  HMDB-51~\cite{HMDB}, and SomethingSomething V2~\cite{ssvth}. Our study covers multiple settings, including zero-shot transfer, few-shot learning, base-to-novel generalization and fully-supervised video recognition.

\noindent
\textbf{Architecture.} We employ the ViCLIP~\cite{InternVid} pre-trained ViT-B/16 in our experiments. For the zero-shot transfer, few-shot learning, base-to-novel generalization and fully-supervised settings, we incorporated our multi-modal spatio-temporal adapter starting from the 1st, 8th, 1st, and 1st transformer blocks and extended it to the last block in both the language and vision components.

\noindent
\textbf{Implementation Details.} Table~\ref{Implementation} outlines the implementation specifics for efficient transfer video recognition. All modules in MSTA are initialized using kaiming initialization~\cite{he2015delving}. We use AdamW~\cite{adamw} as the optimizer, with a weight decay of 0.001 and Adam's \(\beta_1\), \(\beta_2\) set to 0.9 and 0.98, respectively. The linear warm-up consists of 5 epochs. For data augmentation during training, we apply ColorJitter (P=0.8), GrayScale (P=0.2), RandomResizedCrop, and FLIP (flip ratio=0.5). All experiments are conducted on 8 Nvidia Tesla-A100-80G GPUs.

\noindent
\textbf{Evaluation Protocols.} (1) Zero-shot: Following previous works~\cite{XCLIP,OST}, we evaluate zero-shot performance on UCF-101~\cite{UCF101}, HMDB-51~\cite{HMDB}, Kinetics-600~\cite{K600}. In zero-shot setting, we test using single view with 8 frames. (2) Few-shot and base-to-novel generalization: We evaluate on Kinetics-400~\cite{K400}, UCF-101~\cite{UCF101}, HMDB-51~\cite{HMDB}, and SomethingSomething V2~\cite{ssvth}. Following~\cite{VideoMAE}, we adopt a multi-view for evaluation. (3) Fully-supervised: We evaluate the fully-supervised performance on Kinetics-400~\cite{K400}, using 4 clips with 3 crops (i.e. 4$\times$3 views) per video~\cite{XCLIP}. Each view contains 8 sparsely sampled frames.

\noindent
\textbf{Baseline models.} We compare our approach to efficient transfer learning methods~\cite{ER-ZSAR,JigsawNet,XCLIP,A5,Vita-CLIP,ViFiCLIP,MoTE,OST} that based on CLIP for video tasks. Since these methods are specifically designed for CLIP, they cannot be directly applied to ViCLIP. Therefore, we employed common parameter-efficient fine-tuning methods~\cite{Adapter,LoRA,AdaptFormer} on ViCLIP to enable a more comprehensive comparison.

\subsection{Main Results}

\textbf{Base-to-novel generalization.} In Table~\ref{base2new}, we compare our method with the state-of-the-art results under the base-to-novel setting. Our method achieves new state-of-the-art results on all datasets. Compared to OST~\cite{OST} and ActionCLIP~\cite{ActionCLIP}, which model video-specific inductive biases through full fine-tuning, our approach provides better base accuracy with minimal design modifications and demonstrates a significant improvement in novelty accuracy. Our method achieves a better trade-off between base and novel accuracies across all datasets and obtains the overall best harmonic mean. Moreover, it shows enhanced capability in understanding scene dynamics, particularly on temporally challenging datasets like SSv2. Specifically, compared to OST, we improve the recognition accuracy of the novel category on SSv2 from 11.5 to 15.8 (an increase of 30
37\%). Our method also outperforms classic PEFT approaches (i.e., AdaptFormer~\cite{AdaptFormer}, LoRA~\cite{LoRA}), validating the effectiveness of MSTA and consistency constraints in enhancing generalization capabilities.

\begin{table}[t]
\centering
\caption{Comparisons with state-of-the-art methods for zero-shot video recognition on HMDB51, UCF101 and Kinetics-600. We report Top-1 and Top-5 accuracy using single-view inference.}
\resizebox{\linewidth}{!}{
\begin{tabular}{l|c|ccc}
\toprule
Method&Frames&HMDB-51&UCF-101& K600 \\
\midrule
\rowcolor{gray!5}\multicolumn{5}{c}{Zero-shot pre-trained models}\\
ER-ZSAR$_{\textcolor[rgb]{0.5, 0.5, 0.5}{\rm (ICCV'21)}}$&16& 35.3 $\pm$ 4.6& 51.8 $\pm$ 2.9& 42.1 $\pm$ 1.4  \\
JigsawNet$_{\textcolor[rgb]{0.5, 0.5, 0.5}{\rm (ECCV'22)}}$&16& 38.7 $\pm$ 3.7&56.0 $\pm$ 3.1& -\\
\midrule
\rowcolor{yellow!5}\multicolumn{5}{c}{Adapting pre-trained models}\\
Vanilla CLIP$_{\textcolor[rgb]{0.5, 0.5, 0.5}{\rm (ICML'21)}}$  &32&40.8 $\pm$ 0.3& 63.2 $\pm$ 0.2& 59.8 $\pm$ 0.3\\ 
XCLIP$_{\textcolor[rgb]{0.5, 0.5, 0.5}{\rm (ECCV'22)}}$&32&44.6 $\pm$ 5.2& 72.0 $\pm$ 2.3& 65.2 $\pm$ 0.4 \\ 
A5$_{\textcolor[rgb]{0.5, 0.5, 0.5}{\rm (ECCV'22)}}$&8 / 32&44.3 $\pm$ 2.2& 69.3 $\pm$ 4.2& 55.8 $\pm$ 0.7\\ 
Vita-CLIP$_{\textcolor[rgb]{0.5, 0.5, 0.5}{\rm (CVPR'23)}}$&32& 48.6 $\pm$ 0.6& 75.0 $\pm$ 0.6& 67.4 $\pm$ 0.5\\ 
\midrule
\rowcolor{green!5}\multicolumn{5}{c}{Tuning pre-trained models}\\
ViFi-CLIP$_{\textcolor[rgb]{0.5, 0.5, 0.5}{\rm (CVPR'23)}}$& 32
&51.3 $\pm$ 0.7& 76.8 $\pm$ 0.8& 71.2 $\pm$ 1.0\\
OST$_{\textcolor[rgb]{0.5, 0.5, 0.5}{\rm (CVPR'24)}}$&8& 54.9 $\pm$ 1.1 & \uline{77.9} $\pm$ 1.3& 73.9 $\pm$ 0.8\\
ViCLIP$_{\textcolor[rgb]{0.5, 0.5, 0.5}{\rm (ICLR'23)}}$&8& 51.9 $\pm$ 0.7 & 77.1 $\pm$ 0.9 & 71.5  $\pm$ 1.1   \\
\midrule
\rowcolor{blue!5}\multicolumn{5}{c}{Efficient transfer pre-trained models}\\
Zero-ViCLIP$_{\textcolor[rgb]{0.5, 0.5, 0.5}{\rm (ICLR'23)}}$&8
& 39.8 $\pm$ 0.9& 62.6 $\pm$ 0.5 & 59.2 $\pm$ 0.4\\
+ AdaptFormer$_{\textcolor[rgb]{0.5, 0.5, 0.5}{\rm (NeurIPS'22)}}$&8&54.1 $\pm$ 0.9 & 77.2 $\pm$ 0.6  & 72.8 $\pm$ 0.8 \\
+ LoRA$_{\textcolor[rgb]{0.5, 0.5, 0.5}{\rm (NeurIPS'22)}}$&8&54.2 $\pm$ 0.5 & 77.3 $\pm$ 0.8  & 73.8 $\pm$ 0.7 \\
+MSTA&8& \uline{55.3} $\pm$ 0.7 & 77.8 $\pm$ 0.9  & \uline{74.1} $\pm$ 0.7 \\
+MSTA+$\mathcal{L}_{\rm CC}$&8& \textbf{55.8} $\pm$ 0.8 & \textbf{78.7} $\pm$ 0.9  & \textbf{74.5} $\pm$ 0.5 \\
\bottomrule
\end{tabular}}
\label{zeroshot}
\end{table}



\noindent
\textbf{Few-shot learning.} 
In Table~\ref{fewshot}, we conducted a challenging full-shot few-shot video recognition task, which requires rapid adaptation to new categories with limited samples, demanding both specialization and generalization. Our approach achieved 9 out of 12 best performances across different sample settings for all datasets, demonstrating strong learning capability and transferability. Notably, compared to the second-best method, MoTE~\cite{MoTE}, we used only 10\% of its parameter count (i.e., 8.7M vs. 88M).

\noindent
\textbf{Zero-shot transfer.} We present our zero-shot video recognition results in Table~\ref{zeroshot} and compare our approach with current state-of-the-art methods. First, the model is fine-tuned on the Kinetics-400 dataset~\cite{K400} and directly evaluated on downstream datasets to assess its generalization capability to unseen classes. As shown in the table, our method significantly outperforms the conventional uni-modal zero-shot video recognition pipeline~\cite{ER-ZSAR,JigsawNet}. Furthermore, we compare our approach with methods~\cite{XCLIP,A5,Vita-CLIP,OST} that adapt the CLIP model for zero-shot recognition using Kinetics-400, as well as methods that fine-tune ViCLIP with classic PEFT approaches (e.g., AdaptFormer~\cite{AdaptFormer}, LoRA~\cite{LoRA}). We observe consistent improvements across all datasets compared to these methods.

\noindent
\textbf{Fully-supervised learning.} We also conducted fully supervised experiments on the large-scale video benchmark dataset Kinetics-400 to validate the effectiveness of our method in supervised settings. As shown in Table~\ref{fully}, our method achieved the best performance. Specifically, our method outperforms Vita-CLIP~\cite{Vita-CLIP} by approximately 2\% on K400.

\begin{table}[t]
\centering
\caption{Comparisons with state-of-the-art methods for fully-supervised video recognition on Kinetics-400. We report Top-1 and Top-5 accuracy using single-view inference.}
\resizebox{\linewidth}{!}{
\begin{tabular}{l|c|c|cc}
\toprule
\multirow{2}{*}{Method}&\multirow{2}{*}{Frames}&Tunable&\multicolumn{2}{c}{K-400}\\\cline{4-5}
& &Param& Top-1&Top-5\\
\midrule
XCLIP$_{\textcolor[rgb]{0.5, 0.5, 0.5}{\rm (ECCV'22)}}$&8& 132&82.3 &95.8\\ 
Vita-CLIP$_{\textcolor[rgb]{0.5, 0.5, 0.5}{\rm (CVPR'23)}}$&8& 39&80.5& \uline{95.9}\\ 
OST$_{\textcolor[rgb]{0.5, 0.5, 0.5}{\rm (CVPR'24)}}$&8 & 149.6&\uline{82.0}&95.8\\ \midrule
ViCLIP$_{\textcolor[rgb]{0.5, 0.5, 0.5}{\rm (ICLR'23)}}$&8& 124.3&79.9&95.1\\
+ AdaptFormer$_{\textcolor[rgb]{0.5, 0.5, 0.5}{\rm (NeurIPS'22)}}$&8&7.9 &80.2& 95.5\\
+ LoRA$_{\textcolor[rgb]{0.5, 0.5, 0.5}{\rm (NeurIPS'22)}}$&8&9.4 &80.3&95.4\\
+ MSTA&8& 8.7&81.6 & 95.8 \\
+ MSTA+$\mathcal{L}_{\rm CC}$&8& 8.7&\textbf{82.2}&\textbf{96.2}\\
\bottomrule
\end{tabular}}
\label{fully}
\end{table}

\begin{table*}[t]
\caption{Ablation studies over 4 datasets used in base-to-novel generalization setting.}
\label{ablation}
\centering
\begin{subtable}[t]{0.4\textwidth}
\centering
\begin{tabular}{c|ccc}
\toprule
Model Variants & Base & Novel & HM \\
\midrule
Only L-Adapter&66.1 &51.5 &57.9 \\
Only V-Adapter&65.7 &51.7 &57.9 \\
w/o Shared Layers&68.0 & 52.9&59.5\\
Full&\textbf{68.6}&\textbf{53.5}&\textbf{60.1}  \\
\bottomrule
\end{tabular}
\caption{Performance with Different Model Variants}
\label{model_variants}
\end{subtable}
\hfill
\begin{subtable}[t]{0.28\textwidth}
\centering
\begin{tabular}{c|ccc}
\toprule
Dims & Base & Novel & HM \\
\midrule
64 & 67.5 & 52.2& 58.9 \\
128 &68.2&53.1 & 59.7  \\
256 & 68.6 &\textbf{53.5} &\textbf{60.1} \\
512 & \textbf{68.7}& 53.3&60.0 \\
\bottomrule
\end{tabular}
\caption{Dimensions of Shared Layers}
\label{dimension}
\end{subtable}
\hfill
\begin{subtable}[t]{0.28\textwidth}
\centering
\begin{tabular}{c|ccc}
\toprule
$\lambda$ & Base & Novel & HM \\
\midrule
0.001 & 67.7 & 53.1 &59.5\\
0.005 &\textbf{68.6} &\textbf{53.5} &\textbf{60.1} \\
0.01 &68.5  & 52.9&59.7\\
0.05 & 67.1 & 40.3&50.4 \\
\bottomrule
\end{tabular}
\caption{Scaling Factor}
\label{s_factor}
\end{subtable}

\vspace{1em} 

\begin{subtable}[t]{0.25\textwidth}
\centering
\begin{tabular}{c|ccc}
\toprule
$\alpha$ & Base & Novel & HM \\
\midrule
0.1 & 68.1&53.2 &59.7\\
1.0 &\textbf{68.6} &\textbf{53.5} &\textbf{60.1}\\
5.0 & 68.5 & 53.0&59.8\\
\bottomrule
\end{tabular}
\caption{Weighting Factor}
\label{w_factor}
\end{subtable}
\hfill
\begin{subtable}[t]{0.25\textwidth}
\centering
\begin{tabular}{c|ccc}
\toprule
$N$ & Base & Novel & HM \\
\midrule
2 &68.6 &\textbf{53.5 }&\textbf{60.1}\\
4 & \textbf{68.8} & 53.0&59.9\\
8 & 68.6 & 52.9&59.7\\
\bottomrule
\end{tabular}
\caption{Discription Number}
\label{Discription}
\end{subtable}
\hfill
\begin{subtable}[t]{0.44\textwidth}
\centering
\begin{tabular}{c|ccccc}
\toprule
Layer &1$\rightarrow$6&1$\rightarrow$12 &7$\rightarrow$12 &8$\rightarrow$12& 10$\rightarrow$12 \\
\midrule
Base  & 67.4 &68.6&68.1&\textbf{68.7}&65.9 \\
Novel & 52.9 &\textbf{53.5}&53.0&53.1&51.9\\
HM & 59.3 &\textbf{60.1}&59.6&59.9& 58.1 \\
\bottomrule
\end{tabular}
\caption{MSTA inserted layers.}
\label{inserted}
\end{subtable}
\end{table*}

\subsection{Ablation Studies}
We conduct ablation studies on base-to-novel settings in Table~\ref{ablation} to investigate the learning capacity and generalizability of our model in different instantiations.

\noindent
\textbf{Component-wise analysis of MSTA.} We began by examining the impact of various adapter configurations. Specifically, we compared the performance of uni-modal adapters, which are adapters applied exclusively to either the vision (V-) or language (L-) branch, against the performance of our full MSTA model, which incorporates adapters in both modalities. Our findings, as presented in Table 4a, clearly indicate that the uni-modal adapters are outperformed by the full MSTA model. This suggests that the integration of both visual and textual information is crucial for optimal performance. Moreover, we discovered that omitting the shared projection layers (w/o Shared Layers) results in a noticeable performance drop. When these shared layers are included, we observe an increase in the harmonic mean (HM) from 59.5 to 60.1 across the four datasets.

\noindent
\textbf{Dimension of the Shared Layer.} The dimension of the shared layer in our MSTA determines the number of parameters required to extract relationships from the features of the two modalities. We conducted an ablation study by systematically varying the dimension of the shared layer to assess its impact. As shown in Table~\ref{dimension}, accuracy for base categories peaks as the intermediate dimension increases, while accuracy for novel categories saturates around 256. This may be due to larger dimensions introducing more trainable parameters, which increases the risk of over-fitting.

\noindent
\textbf{Scaling Factor $\lambda$.}  
The scaling factor $\lambda$ balances general and task-specific features in MSTA. As shown in Table~\ref{s_factor}, we found $\lambda = 0.005$ achieves the best trade-off (HM) between base and novel categories. Larger values enhance base-category performance but reduce generalization, while smaller values limit adaptability.

\noindent
\textbf{Weighting Factor $\alpha$.} The weighting factor $\alpha$, associated with the consistency constraint loss, is a critical hyperparameter affecting model performance. As demonstrated in Table~\ref{w_factor}, varying $\alpha$ significantly impacts model accuracy, with optimal performance observed at $\alpha = 1.0$. Deviating from this optimal value, either by increasing or decreasing $\alpha$, results in performance degradation. These observations highlight the necessity of precise tuning of the consistency constraint to achieve an effective balance between model adaptability and generalization.

\noindent
\textbf{Numbers of Descriptions.} We investigate the influence of varying the number of descriptions in Table~\ref{Discription}. We conducted experiments with 2, 4, and 8 spatio-temporal descriptions. It can be observed that the performance is better when $N=2$. This may be because as $N$ increases, the hallucination problem of the LLM becomes more severe, resulting in a significant amount of noisy descriptions.

\noindent
\textbf{Variants of Adding MSTA.} We investigate the impact of incorporating MSTA into various encoder layers, with the results summarized in Table~\ref{inserted}. The performance consistently improves as the number of MSTA layers increases. Additionally, for the same number of MSTA layers, placing them in higher encoder layers (farther from the input) yields superior results. For instance, integrating MSTA into layers 7 to 12 achieves an HM score 0.3 points higher than integrating it into layers 1 to 6, despite both configurations involving six MSTA layers. Importantly, inserting MSTA into layers 8 to 12 provides the best performance for base class tasks, highlighting its particular benefit in few-shot learning scenarios that demand rapid adaptation to novel categories.

\section{Conclusion}

In conclusion, this paper introduces a novel Multi-modal Spatio-Temporal Adapter (MSTA) designed for efficient transfer learning in video-language foundation models. The MSTA aims to enhance alignment between text and video representations, achieving an optimal balance between general knowledge and task-specific knowledge. The proposed method incorporates a spatio-temporal description-guided consistency constraint to reduce overfitting and improve generalization, particularly beneficial in scenarios with limited data. Extensive experiments conducted on diverse tasks demonstrate outstanding performance and state-of-the-art results, achieved with minimal trainable parameters. Comprehensive ablation studies further confirm the effectiveness and scalability of the proposed approach. Overall, our work significantly advances the field of video action recognition by offering an efficient transfer learning framework that maintains the generalizability of pre-trained models while effectively adapting them to specific downstream tasks.

{
    \small
    \bibliographystyle{ieeenat_fullname}
    \bibliography{main}
}


\end{document}